\documentclass{article}

\usepackage{arxiv}

\pdfoutput=1
 
\usepackage[utf8]{inputenc} 
\usepackage[T1]{fontenc}    
\usepackage{hyperref}       
\usepackage{url}            
\usepackage{booktabs}       
\usepackage{amsfonts}       
\usepackage{nicefrac}       
\usepackage{microtype}      
\usepackage{cleveref}       
\usepackage{lipsum}         
\usepackage{graphicx}
\usepackage{natbib}
\usepackage{doi}
\usepackage{makecell}
\usepackage[table]{xcolor}

\title{A Case Study for Compliance as Code with Graphs and Language Models: Public release of the Regulatory Knowledge Graph}

\author{ \href{https://orcid.org/0000-0002-4460-4426}{\includegraphics[scale=0.06]{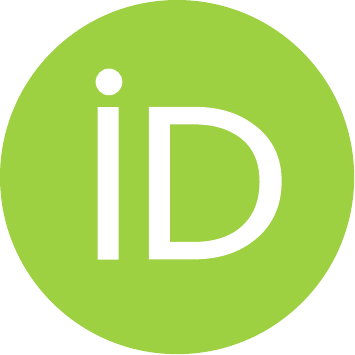}\hspace{1mm}Vladimir D.~Ershov}\thanks{https://www.linkedin.com/in/vladimir-ershov-33559466} \\
	Clausematch Research\\
	London, United Kingdom \\
	\texttt{vladimir.ershov@clausematch.com} \\
}

\renewcommand{\shorttitle}{\textit{arxiv submission}}

\hypersetup{
pdftitle={A template for the arxiv style},
pdfsubject={cs.IR, cs.LG, cs.AI},
pdfauthor={Vladimir D. ~Ershov},
pdfkeywords={First keyword, Second keyword, More},
}

\begin{document}
\maketitle

\begin{abstract}
 The paper presents a study on using language models to automate the construction of executable Knowledge Graph (KG) for compliance. The paper focuses on Abu Dhabi Global Market regulations and taxonomy, involves manual tagging a portion of the regulations, training BERT-based models, which are then applied to the rest of the corpus. Coreference resolution and syntax analysis were used to parse the relationships between the tagged entities and to form KG stored in a Neo4j database. The paper states that the use of machine learning models released by regulators to automate the interpretation of rules is a vital step towards compliance automation, demonstrates the concept querying with Cypher, and states that the produced sub-graphs combined with Graph Neural Networks (GNN) will achieve expandability in judgment automation systems. The graph is open sourced on GitHub to provide structured data for future advancements in the field.
\end{abstract}


\section{Introduction}
Regulatory compliance is a crucial aspect of many industries, and ensuring that organizations abide by relevant regulations is vital for maintaining trust and preventing potential harm. The global financial crisis of 2008 and cases such as Enron and FTX have demonstrated the severe impact on society that can occur when compliance is not enforced. However, ensuring compliance is a complex and time-consuming process that requires a thorough understanding of relevant regulations and their application to specific situations. Given the cross-organizational nature of many businesses and the constant flow of domestic and international regulatory changes, it may not be possible for most of organizations to stay up-to-date on compliance without automation. To address the challenge, this study proposes an innovative approach to compliance automation through the use of pre-trained language models and knowledge graph technology. The paper details the steps involved in the project, including data collection, machine learning (ML) models development, KG construction, and experiment outcomes. Additionally, the study outlines future steps, such as training deep learning models for relation extraction and using GNNs for decision automation.

The study draws on previous research in NLP and GNNs to support the use of language models and graph-based algorithms in compliance decision making. For example, the effectiveness of KG and GNNs in modeling compliance risks and automating decisions based on the model is demonstrated in the paper "Towards knowledge graph reasoning for supply chain risk management using graph neural networks" \cite{kosasih2022towards}. The paper "Large-Scale Multi-Label Text Classification on EU Legislation" \cite{chalkidis2019large} also demonstrates that the use of pre-trained language model even in the legal domain can often be superior to other even custom-made architectures.
\section{Background and related work}
Different fields of compliance have been investigated for more than 30 years and this has resulted in a vast variety of papers and surveys. These include Corporate compliance \cite{mcgreal2009corporate}, Medical compliance \cite{trostle1988medical}, and Business process compliance \cite{fellmann2016state} to name a few. A recent case study illustrates that compliance remains a significant challenge due to excessive, dynamic, and complex requirements that create "impenetrable spaghetti processes" \cite{adams2022banks}.

 One of the most well-researched set of works composed around an abstract formal framework for regulatory compliance (\cite{governatori2006compliance},\cite{governatori2008algorithm},\cite{tosatto2013towards},\cite{governatori2013business},\cite{hashmi2014modeling},\cite{sadiq2015managing},\cite{hashmi2016normative}) covered by \href{https://orcid.org/0000-0002-4460-4426}{Guido Governatori} establish a conceptually sound formalisation of the norms and compliance rules to describe different deontic modalities: obligations, permissions etc. The proof-of-concept prototype presented by \cite{fdhila2022verifying} shows that once the taxonomy of entities and relations forming regulated activities and rules is defined and the formal framework is built, it is possible to use First Order Logic over it. This makes it possible to develop and apply algorithms for compliance verification over formal models. However, the process of rules transition into a formal form is manual, as is the process of developing algorithms to verify a specific type of compliance. The current paper aims to address this challenge by introducing an approach to automate the construction of formalized expressions under specified taxonomy and suggests the means to automate the construction of compliance verification algorithms as well.
 
The challenge faced by state-of-the-art ML systems is their lack of explainability, where predictions are made without clear explanation  \cite{baryannis2019supply}. This is a significant issue in the field of compliance, where explainability is a crucial focus. The Explainability-by-Design Methodology \cite{huynh2022explainability} offers a solution by manually augmenting rules in the decision-making system to provide comprehensive explanations. It involves building a system that organizes regulatory requirements in a graph-based RDF system. However, the research also notes that the cost of manually adding structure for explainability can take a month or more for a single use-case. Therefore, a suitable system for compliance automation at scale should be able to learn and produce explanations without the need for manually labelled data after training is complete. Many studies have been focused on this by automating the rule extraction process in the form of a KG and providing rule-guided explainable decisions based on induced rules \cite{ma2019jointly}. This paper presents an approach to compliance automation by extracting structure from existing regulatory rules and allowing for the general formalisation of rules to emerge, which can be used as explainable input and output for decision automation. Unlike previous works, where formal frameworks were constructed first and existing rules were then converted to a predefined form, this approach does not rely on a manually constructed algorithm.

\section{Method}
\begin{eqnarray}
	\label{eqn:scenario}
	&\verb|If an 'ENT' with this 'PERM' was doing this 'ACT' or 'FS' with 'PROD'| \nonumber
	\\
	&\verb|using 'TECH' then to avoid this 'RISK' they should 'MIT'.|
\end{eqnarray}

\label{sec:headings}
The purpose of presented approach is to explore an applicability of recent advancements in NLP  \citep{bert, transformers, ulmft} for the problem of compliance automatisation. 
From subject matter experts point of view, a core feature of such systems is an ability to extract obligations which in the most general case come in the form scenario (\ref{eqn:scenario}). Given the vital nature of taxonomy for forming the rules and a current form of regulation distribution through regulatory documents suggested course of actions was based on a \textit{Gaia knowledge extraction system} \cite{li2020gaia} :
\begin{enumerate}
	\item To define type of entities generalised from Taxonomy and used by regulators to form rules and obligations
	\item To manually tag these entities in the current regulatory documents to produce a dataset of sufficient size
	\item To apply state of the art approach for Named Entity Recognition (NER) task to develop models capable of tagging these entity types in the unstructured regulatory texts
	\item To manually evaluate model performance to account for concept drift as during labelling expert may change the style of tagging
	\item To assess an applicability of the approach and fine tune models if needed 
	\item To resolve co-references in the documents and apply tagging models to form the nodes of the graph
	\item To extract relations between tagged entities with syntax analysis and combine it with document hierarchy to form edges of the graph 
	\item To explore the outcomes and draft the future steps
\end{enumerate}

\subsection{Dataset}
As the most general and common scenario legal experts suggested to consider scenario (\ref{eqn:scenario}). Given that and under the assumption that entities of these types are participate in forming most of the obligations the taxonomy of entities types listed below was formed. It is listed here in way as it was described to the labelling team.

\begin{itemize}
	\item \textbf{Permissions [PERM]} - Any permission
	\item \textbf{Definitions [DEF]} - Any of the terms defined in the predefined glossary
	\item \textbf{Risks [RISK]} - Any mention of an identified risk, liability or issue. Specifically: if it’s something or a situation that regulated entities need to avoid/be aware of, if they need to have something in place to stop something, if it's something that they need to comply with or compare themselves against
	\item \textbf{Mitigation [MIT]} - Any mention of rule, requirement, or guidance. Specifically: if it’s a rule, if it’s a link to another rule, an expectation, a requirement, a process to follow, an information to include or what should be considered
	\item \textbf{Entities [ENT]} - Any mention of a firm, financial institution or authorised person 
	\item \textbf{Activities [ACT]} - Any concept that pertains to an entity initiating a FS related action/activity
	\item \textbf{A specific FS Concept	[FS]} - Any mention of a financial concept or service such as liquidity, debt, moving money, custody, financing etc.
	\item \textbf{A financial services product [PROD]} - Any mention of a financial product or 'vehicle' holding money
	\item \textbf{Any mention of tech [TECH] } - Any mention of tech i.e Digital Assets, Robo Advisory, AI, Wallets, Encryption, Crypto, DLT,  APIs, Software packages, Accounting packages, Cloud, Data etc.
\end{itemize}

After approximately 6 man-months of team efforts the dataset described in the Table \ref{tab:table1} bellow was labelled. It was done over 1880 paragraphs in the Conduct of Business Rulebook (COBS) document.

\begin{table}[ht]
	\caption{Manually labelled Taxonomy entities}
	\centering
	\begin{tabular}{p{0.25\linewidth} l p{0.15\linewidth} l p{0.1\linewidth}}
		\toprule
		Concept&Tag&Labeled paragraphs requested&Subject matter understanding required   &Entities labelled to train \\ \hline
		\midrule
		Permissions&PERM&200&Easy&89 \\
		Definitions&DEF&200&Easy&2896 \\
		Risks&RISK&1000 to 10000&Hard&2170 \\
		Mitigation&MIT&1000 to 10000&Hard&3298 \\
		Entities&ENT&1000&Medium&2748 \\
		Activities&ACT&200&Easy&1444 \\
		A specific FS Concept&FS&1000&Medium&1404 \\
		A financial services product&PROD&1000&Medium&239 \\
		Any mention of technology&TECH&1000&Medium&257 \\
	\end{tabular}
	\label{tab:table1}
\end{table}

\subsection{Architecture}
\begin{figure}[!htb]
	\centering
	\caption{Clausematch in-text tagging feature used for document labelling}	\label{fig:fig1}	
	\fbox{\includegraphics[scale=0.4]{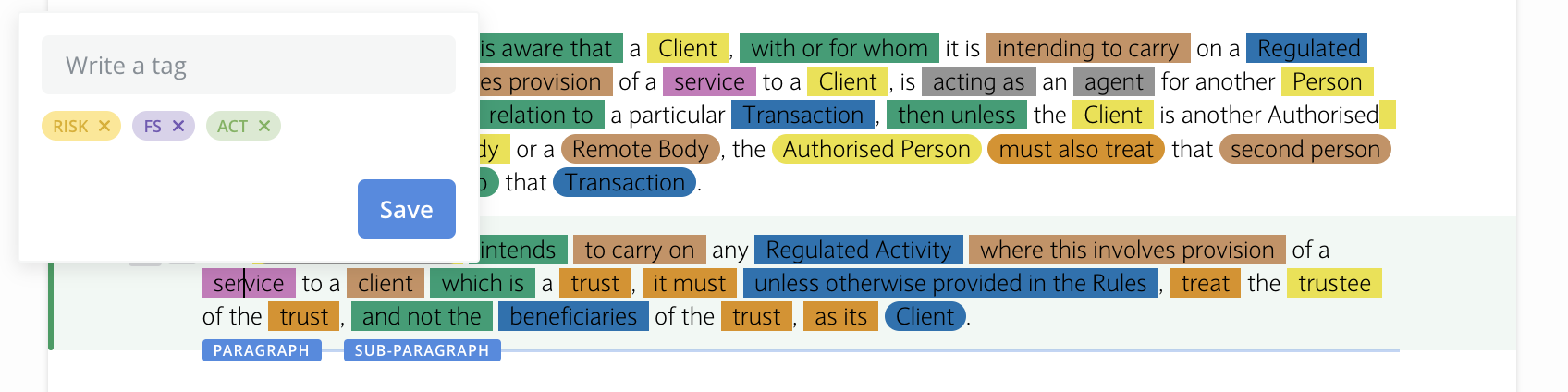}}
\end{figure}

Here is the list of key components, models and libraries used to apply the advancements in NLP to the paper's task:

\paragraph{Labelling} 
Initially labelling was started at Clausematch environment in the wiki style markup. Clausematch is a SAAS solution which is used for compliance and provides document storage and drafting functionality. It gives an advantage for expert to access content of the whole document and have a real-time collaboration in the browser. Disadvantage was that wiki markup requires human to type set of additional characters, which due to various inconsistencies and typos would damage up to 10\% of paragraphs. Thus the in-text tagging feature shown in Figure \ref{fig:fig1} was introduced to eliminate that. \textit{Doccano} \cite{Doccano} was used for a manual evaluation and labelling datasets for the fine-tuning. In comparison to Clausematch in-text tagging doccano's key advantages are an extensive set of hot keys to support the labelling work and UI with a real-time statistic on the tagging progress. Disadvantage is that each paragraph is detached from the document and an overall context of the clause is not available for the expert.

\paragraph{Applied machine learning tools} 
For the development environment \textit{Python 3.7.3} and \textit{spaCy 2.2} \cite{spacy2} were used. The author used the  \textit{XLNet en\_trf\_xlnetbasecased\_lg}  \cite{yang2019xlnet} variation of pre-trained language model as a base and the NER pipe in the spaCy framework was updated during the training by applying the BILUO scheme. \textit{Neuralcoref} \cite{neuralcoref} spaCy extension was used for coreference resolution. For the construction of graph edges presenting relations between entities in-text occurrences non-monotonic arc-eager transition-system  \cite{honnibal-johnson-2015-improved} with a custom sentence segmentation \cite{nivre-nilsson-2005-pseudo} was used, which is available as a part of spaCy pipeline. 

\paragraph{Infrastructure and experiment tracking}
AWS cloud was used as a GPU computation and storage resource provider. The experiment tracking was organised with \textit{MLflow} \cite{zaharia2018accelerating} and S3 for the metadata storage. S3 was also used to store models, inputs and outputs for training and evaluation pipelines. These artifacts were indexed with a data version control (DVC)\footnote{https://dvc.org} tool. 

\paragraph{Visualisation and graph exploration}
A key expectation for the applicability of new technologies is to provide an access for humans to the produced results which is particularly important in the field of governance and compliance. The Clausematch UI, as it shown in Figure \ref{fig:fig1}, provides access to the entities tagged by the models. For a deep and comprehensive analysis of the graph structure, the author used \textit{Neo4j} \cite{ozgur2018comparative} as it supports \textit{Cypher} \cite{francis2018cypher} and the \textit{Bloom} \cite{hodlergraph} tool. The next section will cover the results and present the visuals.
	
\section{Results}
The developed system automates the process of extracting KG from compliance documents. This includes automating the training of tagging models and running these models against text on the Clausematch platform. \textbf{The outcome of the automated data structuring in the form of a Neo4j dump we release publicly within these paper and supplementary visuals on the} \textit{GitHub}\footnote{https://github.com/Vladimir-Ershov/adgm-kg1}. The following sections will provide more details on the models used to produce the data and the dataset itself.

\begin{table}[ht]
	\caption{Grouping and manual evaluation results}
	\centering
	\begin{tabular}{ l l l l l l}
			\toprule
			Tag& Dataset length&Precision&Recall&F1 & Model \\ \hline
			\midrule
			PERM&166&91.01&96.43& 93.64 & PERM\\
			RISK&255&86.24&80.50&83.27 & RISK\_MIT \\
			MIT&255 &81.54&55.58&66.11 & RISK\_MIT\\
			ENT&127&96.57&97.40&96.98 & ENT\\
			ACT&180&91.03&91.67&91.35 & ACT\_FS\_PROD \\
			FS&180&94.97&79.41&86.50 & ACT\_FS\_PROD \\
			PROD&180&90.32&93.33&91.80 & ACT\_FS\_PROD \\
		\end{tabular}
		\label{tab:table2}
	\end{table}

\paragraph{Models}
\label{sec:models}
The process of training the models presents some challenges. It appears to not be possible to train a single XLNet-based model to capture all of the tags at once. As a result, the tags were organized into groups, and a separate model was trained and manually evaluated for each group. The results of the grouping and evaluation are presented in Table \ref{tab:table2}. Another challenge is consistency in tagging, as experts tend to reshape their perception of the meaning of taxonomy entities during the entity tagging process. For that reason, the evaluation in Table \ref{tab:table2} was not done on the dev dataset, but instead, manual evaluation was used. The empirical perception is that up to one out of four entities may be tagged differently by the same expert after several hundred paragraphs. Since the  \textit{DEF} tag presents definitions listed in the glossary, a simplified version for tagging was used as a combination of extracted UPOS combined with lemma as a mask to match in the text.

\paragraph{Tags}
The models described above were used to extract KG containing:
\begin{itemize}
	\item Relationships: 1,209,207 
	\item Nodes: 231,404 
	\item "TagOccur" nodes: 173853 - which means occurrences of the tags in text
	\item "Tag" nodes: 35498 - for amount of entities divided into different concept. Detailed data listed in the \ref{sec:appx1}
	\item "Document" nodes: 26
	\item "Paragraph" nodes: 22027
\end{itemize} 
Tags allow for a high-level overview of the statistics of discovered entities. Appendix \ref{sec:app_pie} demonstrates one way to visualize it according to the detected Product in different documents. The corresponding bar plots and heatmap for the most popular tag co-occurrences can be found on \textit{GitHub}\footnote{https://github.com/Vladimir-Ershov/adgm-kg1}.

\begin{figure}[htb!]
	\centering
	\caption{Two documents(green) intersected by entities identified as Product(blue)  in paragraphs(tawny) by model. Neo4j interactive visualization formed with a Cypher query \ref{sec:intersect_prod} }	\label{fig:intersect_prod}	
	\fbox{\includegraphics[scale=0.28]{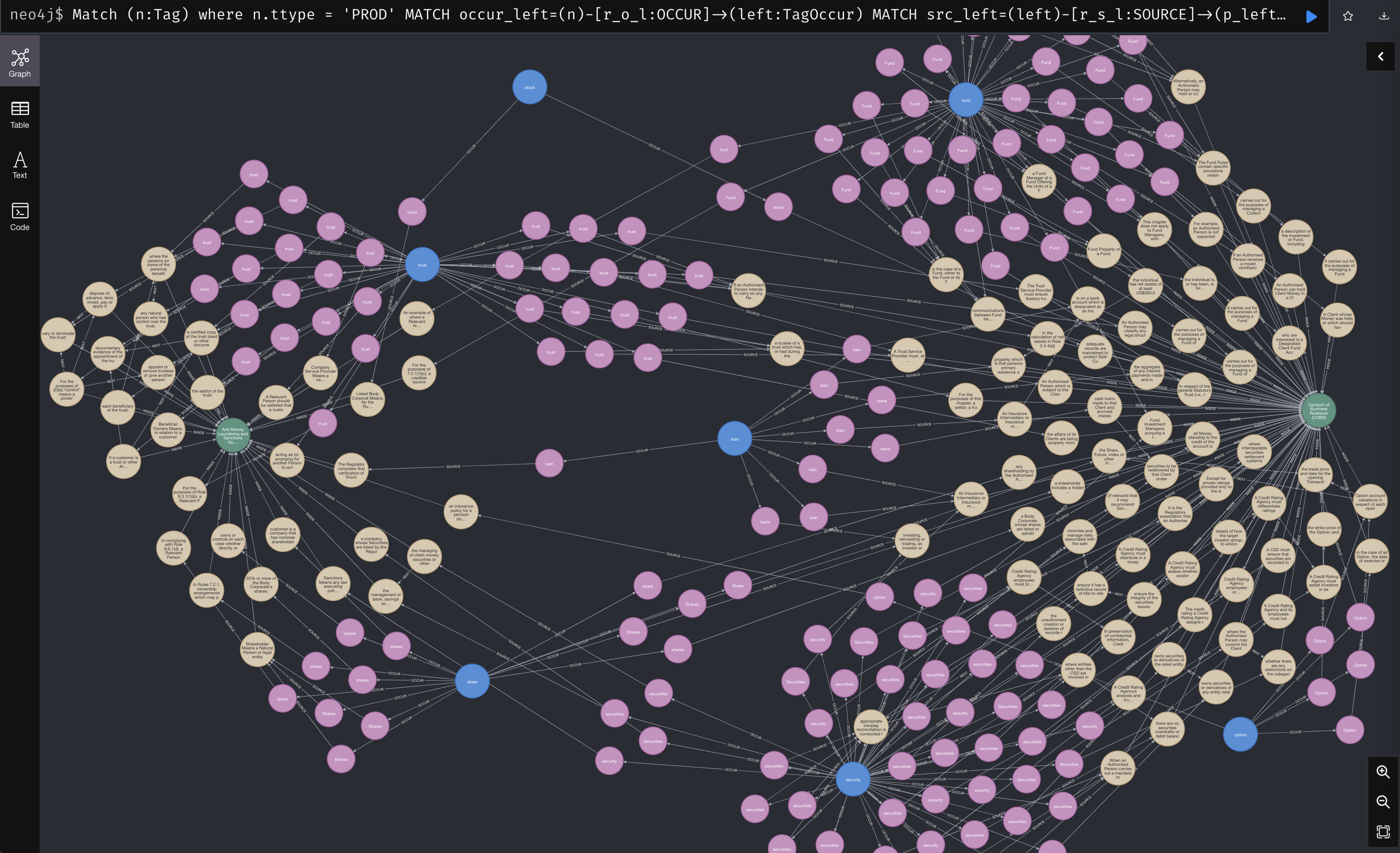}}
\end{figure}

\paragraph{Graph}
As parent and child relationships for the paragraphs in documents are part of the graph, the content can be explored using the interactive Neo4j tool, starting from the table of contents, using \textit{Cypher} as demonstrated in Appendix  \ref{sec:neo1}. After discussion with regulatory compliance experts, the following visual was included as \textbf{evidence of progress towards compliance automation: documents can now be analyzed on the concept level} using query \ref{sec:intersect_prod}, as shown in Figure \ref{fig:intersect_prod}.
 
 The Neo4j database allows querying the data through \textit{REST} in the \textit{JSON} format. This means that the regulations provided in the Regulatory Knowledge Graph, along with the results of extracted entities and relations, are available for integration into compliance automation systems as a source of ground truth. The graph structure allows for capturing and propagating relations in a new dimension of concepts, making it possible for a downstream system to replicate an associative context expansion. As an example, Appendix \ref{fig:path} shows an application of graph algorithms to the extracted structure, capturing all the shortest paths of length 4 between entities related to \textit{insurance} and \textit{rule}. For visualization purposes, the Bloom tool was also used: Appendix \ref{sec:bloom}. Bloom allows for easy handling of graphs with 1000 or more nodes and better customization, such as conditional colour coding of the nodes.

\paragraph{Principal finding}
One of the key finding of this paper is a proposal that fine-tuned language models applied to the text for tagging and relation extraction produce interpretation in the form of layer outputs and therefore capture the meaning of a concept from the training data. These embeddings introduce dimensions to automate the formalisation of statements' juxtapositions from the text, making it possible to introduce explainability in the form of a sub-graph and causal inference for automated decision-making based on the knowledge graph (KG). For that all nodes and relations in the graph derived from text have to be a result of unified interpretation. Therefore entities and relations between them have to be extracted according to the taxonomy implied by experts for compliance rule interpretation and learned by ML models. The consequence is that \textbf{the distribution of ML models containing interpretations of the rules within regulations by regulators is a vital step towards compliance automation} and will introduce a common ground where match between obligations and business internal processes is possible. Regulated companies will be able to apply released models to their internal Policies and Controls and produce KG as it would be seen by regulator. The internal KG and Regulatory KG could serve as input for GNN or other form of ML model to automate all range of compliance tasks: gap analysis, contradiction detection, compliance verification, etc. The interpretation of the rules introduced by the same ML models will allow the system to highlight factual differences between documents, rather than differences in interpretation. 
\begin{table}[htb!]
	\caption{The most common relations between extracted entities as it is expected by the expert. Single cell reads as 'PERM'-[Allow]->'ACT', 'PERM'-[Authorise]->'ACT', 'PERM'-[Involving]->'ACT'}
	\centering
	\scalebox{0.7} {		
		\begin{tabular}{ l l l l l l l l l l}
			\toprule
			Tag&PERM&ACT&DEF&RISK&MIT&ENT&PROD&FS&TECH\\ \hline
			\midrule
			PERM&\cellcolor{lightgray!25}&\makecell{Allow\\Authorise\\Involving}&\makecell{Involving\\Relating \\Uses}&\makecell{Create\\Increase\\Decreases}&\makecell{Must ensure\\Decreases}&\makecell{Involving\\Relating \\Uses}&\makecell{Involving\\Relating \\Uses}&\makecell{Involving\\Relating \\Uses}&\makecell{Involving\\Relating \\Uses}\\
			ACT&\makecell{Involving\\Relating \\Uses}&\cellcolor{lightgray!25}&\makecell{Involving\\Relating \\Uses}&\makecell{Create\\Increase\\Decreases}&\makecell{Must ensure\\Decreases}&\makecell{Involving\\Relating \\Uses}&\makecell{Involving\\Relating \\Uses}&\makecell{Involving\\Relating \\Uses}&\makecell{Involving\\Relating \\Uses}\\
			DEF&\makecell{Involving\\Relating \\Uses}&\makecell{Involving\\Relating \\Uses}&\cellcolor{lightgray!25}&\makecell{Create\\Increase\\Decreases}&\makecell{Must ensure\\Decreases}&\makecell{Involving\\Relating \\Uses}&\makecell{Involving\\Relating \\Uses}&\makecell{Involving\\Relating \\Uses}&\makecell{Involving\\Relating \\Uses}\\
			RISK&\makecell{Impact\\Create\\Increase\\Decreases}&\makecell{Impact\\Create\\Increase\\Decreases}&\makecell{Impact\\Create\\Increase\\Decreases}&\cellcolor{lightgray!25}&\makecell{Must ensure\\Decreases}&\makecell{Impact\\Create\\Increase\\Decreases}&\makecell{Impact\\Create\\Increase\\Decreases}&\makecell{Impact\\Create\\Increase\\Decreases}&\makecell{Impact\\Create\\Increase\\Decreases}\\
			MIT&\makecell{Must ensure\\Decreases}&\makecell{Must ensure\\Decreases}&\makecell{Must ensure\\Decreases}&\makecell{Create\\Increase\\Decreases}&\cellcolor{lightgray!25}&\makecell{Must ensure\\Decreases}&\makecell{Must ensure\\Decreases}&\makecell{Must ensure\\Decreases}&\makecell{Must ensure\\Decreases}  \\
			ENT&\makecell{Allow\\Authorise\\Cannot\\Involving}&\makecell{Allow\\Authorise\\Cannot\\Involving}&\makecell{Allow\\Authorise\\Cannot\\Involving}&\makecell{Create\\Increase\\Decreases}&\makecell{Must ensure\\Decreases}&\cellcolor{lightgray!25}&\makecell{Manage\\Controlled\\ Owned\\Sell\\Buys}&\makecell{Manage\\Controlled\\Owned\\Sell\\Buys}&\makecell{Manage\\Controlled\\Owned\\Sell\\Buys}\\
			PROD&\makecell{Allow\\Authorise\\Cannot\\Involving}&\makecell{Allow\\Authorise\\Cannot\\Involving}&\makecell{Allow\\Authorise\\Cannot\\Involving}&\makecell{Create\\Increase\\Decreases}&\makecell{Must ensure\\Decreases}&\makecell{Manage\\Controlled\\Owned\\Sell\\Buys}&\cellcolor{lightgray!25}&\makecell{Manage\\Controlled\\Owned\\Sell\\Buys}&\makecell{Involving\\Relating \\Uses}\\
			FS&\makecell{Allow\\Authorise\\Cannot\\Involving}&\makecell{Allow\\Authorise\\Cannot\\Involving}&\makecell{Allow\\Authorise\\Cannot\\Involving}&\makecell{Create\\Increase\\Decreases}&\makecell{Must ensure\\Decreases}&\makecell{Involving\\Relating \\Uses}&\makecell{Involving\\Relating \\Uses}&\cellcolor{lightgray!25}&\makecell{Involving\\Relating \\Uses}\\
			TECH&\makecell{Involving\\Relating \\Uses}&\makecell{Involving\\Relating \\Uses}&\makecell{Involving\\Relating \\Uses}&\makecell{Create\\Increase\\Decreases}&\makecell{Must ensure\\Decreases}&\makecell{Manage\\Controlled\\Owned\\Sell\\Buys}&\makecell{Manage\\Controlled\\Owned\\Sell\\Buys}&\makecell{Manage\\Controlled\\Owned\\Sell\\Buys}&\cellcolor{lightgray!25}\\
		\end{tabular}
	}
	\label{tab:table3}
\end{table}

For the general formalization of rules to emerge, the relation extraction taxonomy must contain a unidirectional \textit{"Is a"} relationship. This is likely to be necessary for a proper resolution of the NEL challenge. The other finding is the proposal on how to approach compliance verification. This challenge is known to have a NP-complexity \cite{tosatto2014business}. The proposal is that KG has to store generalised form of the rules to enable reinforcement learning agents to navigate over sub-graphs of KG \cite{xiong2017deeppath,li2018path} iterating mostly over general form of regulated rules. The key part of the solution here is the ability to make statements juxtaposition possible and to induce precedence order over implied rules. Therefore the relation taxonomy has to be extended with \textit{"precedence"} relationship. Extracted relations between tagged entities at the current state of the graph were derived from text, but there were no classification applied to them.  Table \ref{tab:table3} presents current view of the relation taxonomy formed in discussions with compliance experts as a view on connections needed to form and impose obligation. As previously stated, additional relations suggested to extend this list include: \textit{"Is a"} and \textit{"precedence"}.

\section{Limitations and Future Work}
\begin{figure}[!htb]
	\begin{center}
		\caption{Two different ways to decide on the MIT tag boundaries. Subject matter expert during model evaluation have to set the MIT\_MAN tag. A more granular approach presented on the left was used.}		
		\label{fig:mit}
		\centering
		\begin{minipage}{0.5\textwidth}
			\centering
			\includegraphics[width=0.95\linewidth]{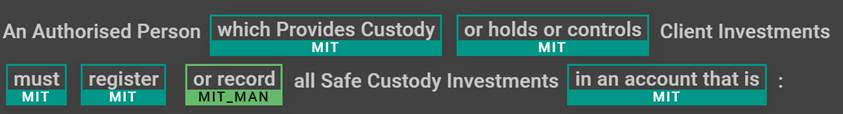}
		\end{minipage}\hfill
		\begin{minipage}{0.5\textwidth}
			\centering
			\includegraphics[width=0.95\linewidth]{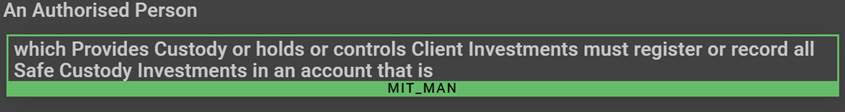}
		\end{minipage}\hfill		
	\end{center}
\end{figure}

\paragraph{RISK\_MIT model}
This case study was focused on Abu Dhabi Global Market (ADGM) regulations and while the approach is expected to be applicable for other regulators, extracted taxonomy, KG extraction pipeline and models may require adjustments. Tagging models might be fine-tuned on a new taxonomy, but it likely to come with a cost of performance on ADGM taxonomy. \verb+XLNet+ baseline model might not be the most efficient to use as training on \verb|RISK| and \verb|MIT| tags didn't provide consistent result. The issue to highlight is that \verb|RISK_MIT| model shows poor results: for some of the regulatory documents on a different topics the performance drops down to 43\%. The reason for that is the same as with the initial model unified for all tags - the generalised concept is too vague and either model size is not capable to handle it or model weights are unlikely to converge. In order to overcome it \verb|RISK| and \verb|MIT| concepts have to be divided into set of sub-concepts per each and separated models should be trained. The other approach might be to train a large language models but that will likely to rise concerns regarding computational efficiency and carbon footprint \cite{patterson2021carbon}. 
\paragraph{Labelling and bias}
During tagging process experts have to decide on the interpretation for tagging boundaries as it shown at Figure \ref{fig:mit}, solving this may not eliminate the challenge and tagging might not be consistent even for a single person. The other challenge is to prepare the dataset diverse enough to reduce potential bias. The data set used for models training was compound from 1880 paragraphs of a single document. That is expected to introduce tagging artifacts which are listed in the Appendix \ref{sec:app_artif}.
\paragraph{Relation extraction and GNN}
The future work for this study includes developing relation extraction models based on the taxonomy from Table  \ref{tab:table3}. Transformer-based architectures, as presented in \cite{yao2019docred}, are likely to be a reliable approach for this task. These models will be used to refine the knowledge graph and make it suitable for use as an input for a compliance automation system. GNNs \cite{scarselli2008graph} can be used as a baseline for such a system, but it is likely that Stepwise Reasoning Networks \cite{qiu2020stepwise} will be more suitable for this task. The use of graph structures is motivated in addition by the NP-complexity of the regulatory compliance task \cite{tosatto2014business}. This is likely to drive the use of reinforcement learning agents that reason over sub-graphs of the KG, which can serve as a Cognitive Database in the context of Cognitive Architecture\cite{kolonin2022cognitive}. This approach may result in "eventual compliance" rather than full compliance.
\paragraph{Graph refinement and NEL}
The other challenge to address is the overall refinement of the graph. Since the extraction of the entities was done on the paragraph level, coreference resolution was used to enrich the text of the paragraph based on the surrounding context. The tagged text in the normalised form of lemmas was used for NEL to generalise different occurrences of the tag under single \verb|Tag| type in the graph. However, this unsupervised process introduced issues that could be addressed in future work.

\section{Discussion and Conclusion}
In conclusion, this study proposes an innovative approach to compliance automation by leveraging the capabilities of language modelling and KG technology. Through the application of ML models and manual annotation of a subset of regulations, the study demonstrates the feasibility of an executable KG that captures interpretation of regulatory taxonomy over regulations. The open-sourced KG serves as a valuable resource for future advancements in the field, and the author suggest the next step of developing automated relation extraction based on the proposed taxonomy. Additionally, the study proposes a call to action for regulators to release ML models to facilitate the capture and distribution of the meaning of regulatory taxonomy and relationships, thus enabling a more efficient and consistent interpretation of regulatory rules. Ultimately, the proposed approach aims to introduce causal inference and explainability in compliance decision-making systems through the use of the Knowledge Graphs.

\section{Acknowledgments}
The author of this study would like to extend his deepest gratitude to the following individuals and organizations, without whom the successful completion of this research would not have been possible.
\begin{itemize}
	\item To Clausematch company, particularly Evgeny Likhoded and Anastasia Dokuchaeva, for their unwavering support throughout the project and commitment to solve compliance once and for all.
	\item To the ADGM, particularly Barry West, for providing the regulations, taxonomy, and subject matter expertise used in this study, as well as valuable feedback and insights.
	\item To colleagues at Clausematch, particularly Christoph Rahmede, who helped to refine the text and style of the paper.
\end{itemize}
We are honoured to have had the opportunity to work with such esteemed individuals and organizations. We are deeply grateful for their support and contributions, without which this research would not have been possible:
\setcitestyle{numbers}
\bibliographystyle{abbrv}
\bibliography{references}
\pagebreak
\appendix
\section{Counts for different tags per concept}
\label{sec:appx1}
\begin{table}[tbh!]
	\centering
	\begin{tabular}{ l l}
			\toprule
			(Tag concept)& counts  \\ \hline
			\midrule
				MIT&20489 \\
				RISK&10737 \\
				TECH&1962 \\
				ACT&654 \\
				FS&583 \\
				ENT&526 \\
				PERM&272 \\
				DEF&202 \\
				PROD&73 \\
	\end{tabular}
\end{table}
\section{Proportions of the products mentions extracted from documents}
\label{sec:app_pie}
\begin{figure}[!htb]
	\begin{minipage}{0.33\textwidth}
		\centering
		\includegraphics[width=1\linewidth]{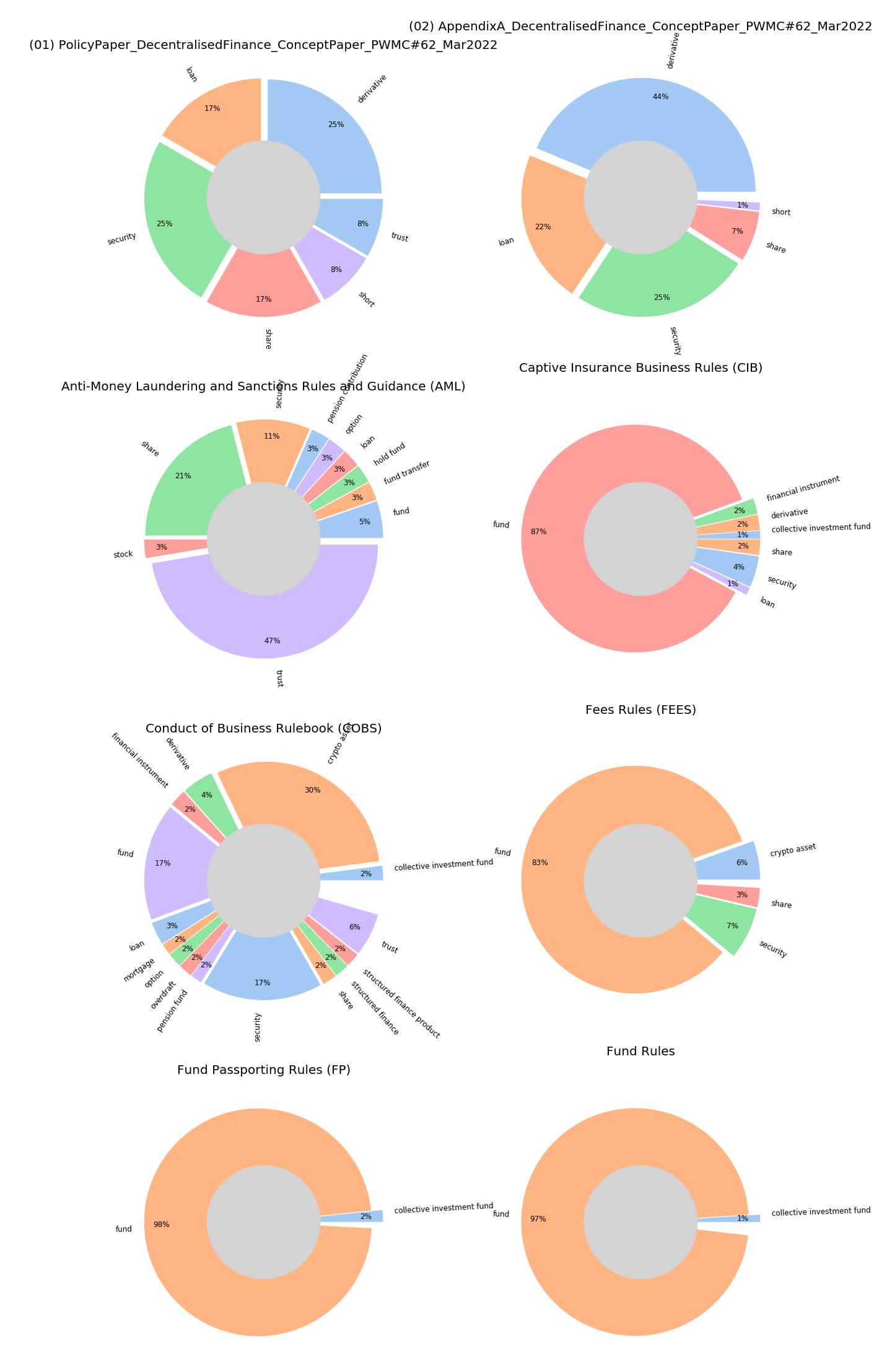}
	\end{minipage}\hfill
	\begin{minipage}{0.34\textwidth}
		\centering
		\includegraphics[width=1\linewidth]{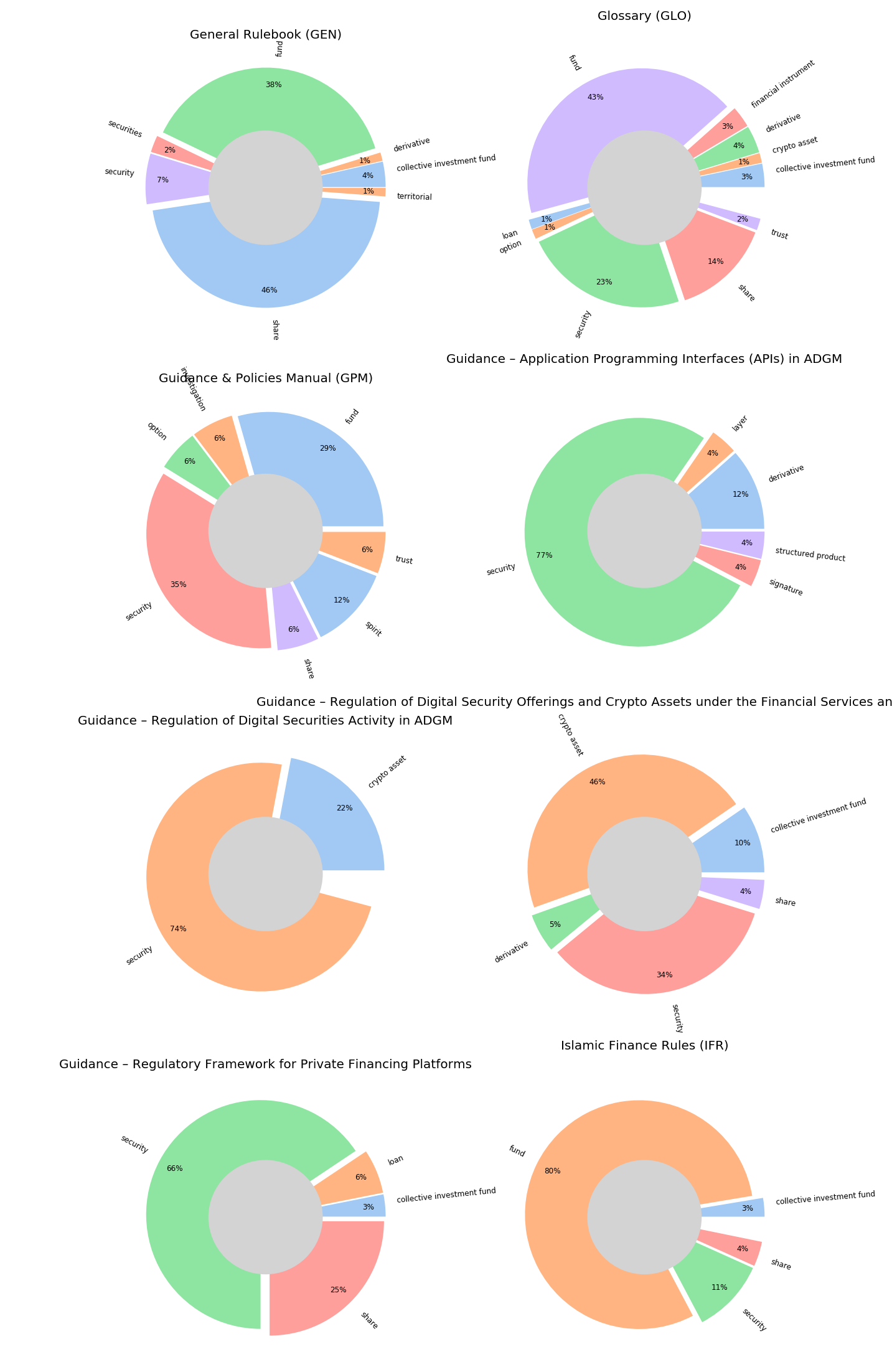}
	\end{minipage}\hfill		
	\begin{minipage}{0.33\textwidth}
		\centering
		\includegraphics[width=1\linewidth]{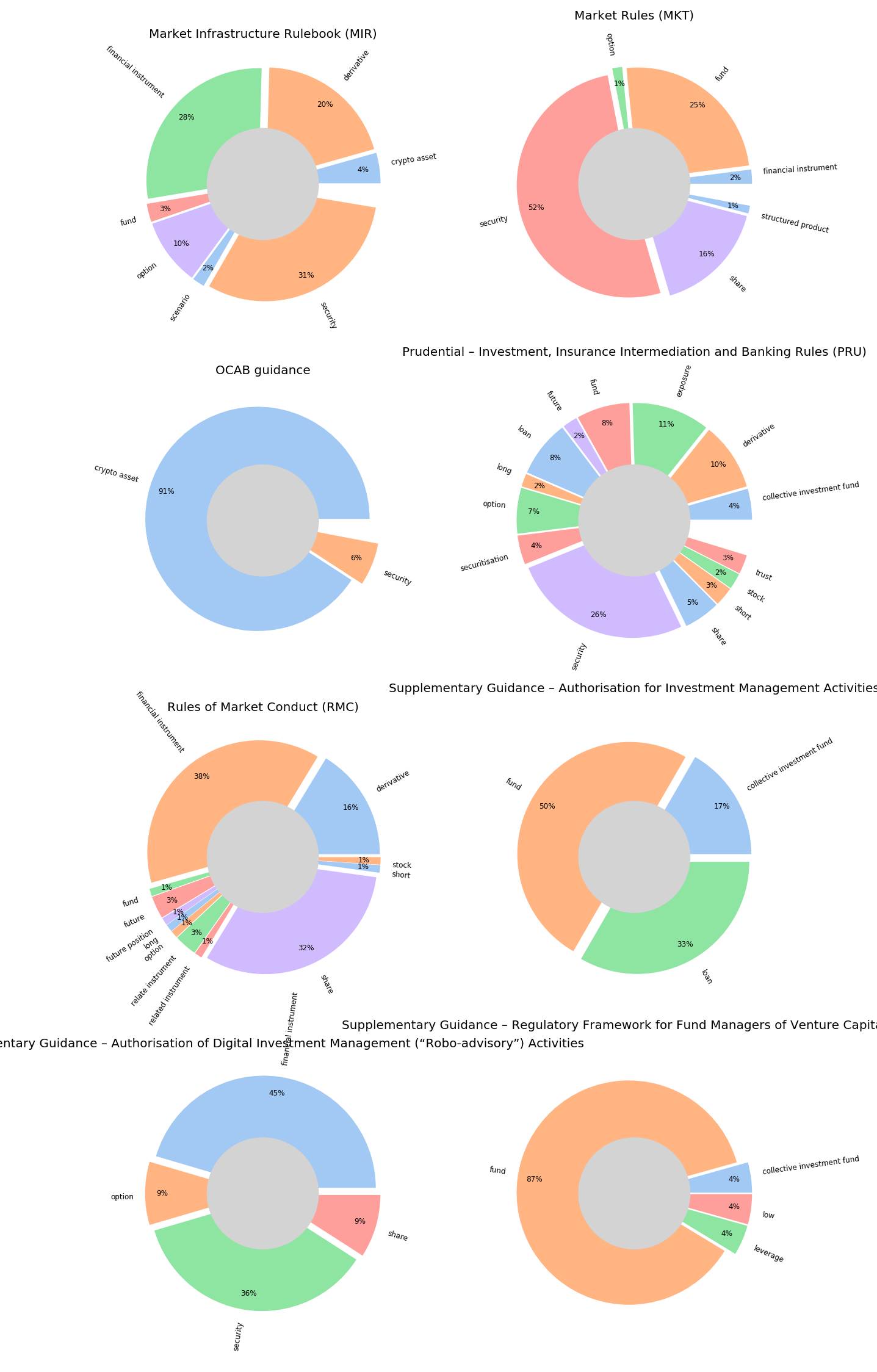}
	\end{minipage}
\end{figure}
\section{Cypher query for intersecting two documents by entities identified as Product}
\label{sec:intersect_prod}
\begin{verbatim}
	Match (n:Tag)
	where n.ttype = 'PROD'
	MATCH occur_left=(n)-[r_o_l:OCCUR]->(left:TagOccur)
	MATCH src_left=(left)-[r_s_l:SOURCE]->(p_left)--(d_left:Document)
	where d_left.title contains '(COBS)'
	MATCH occur_right=(n)-[r_o_r:OCCUR]->(right:TagOccur)
	MATCH src_right=(right)-[r_s_r:SOURCE]->(p_right)--(d_right:Document)
	WHERE d_right.title contains '(AML)'
	RETURN occur_left, occur_right, src_right, src_left,d_left, d_right LIMIT 25000
\end{verbatim}
\section{Getting table of content for the document in Neo4j. The interactive tool allows to expand each node to get internal content}
\label{sec:neo1}
\begin{verbatim}
MATCH p=(f)-[r:NEXT]->(t)--(d:Document)
WHERE f.plevel < 1 and t.plevel < 1 and d.title contains 'AML'	
\end{verbatim}
\begin{figure}[htb!]
	\centering
	\includegraphics[scale=0.15]{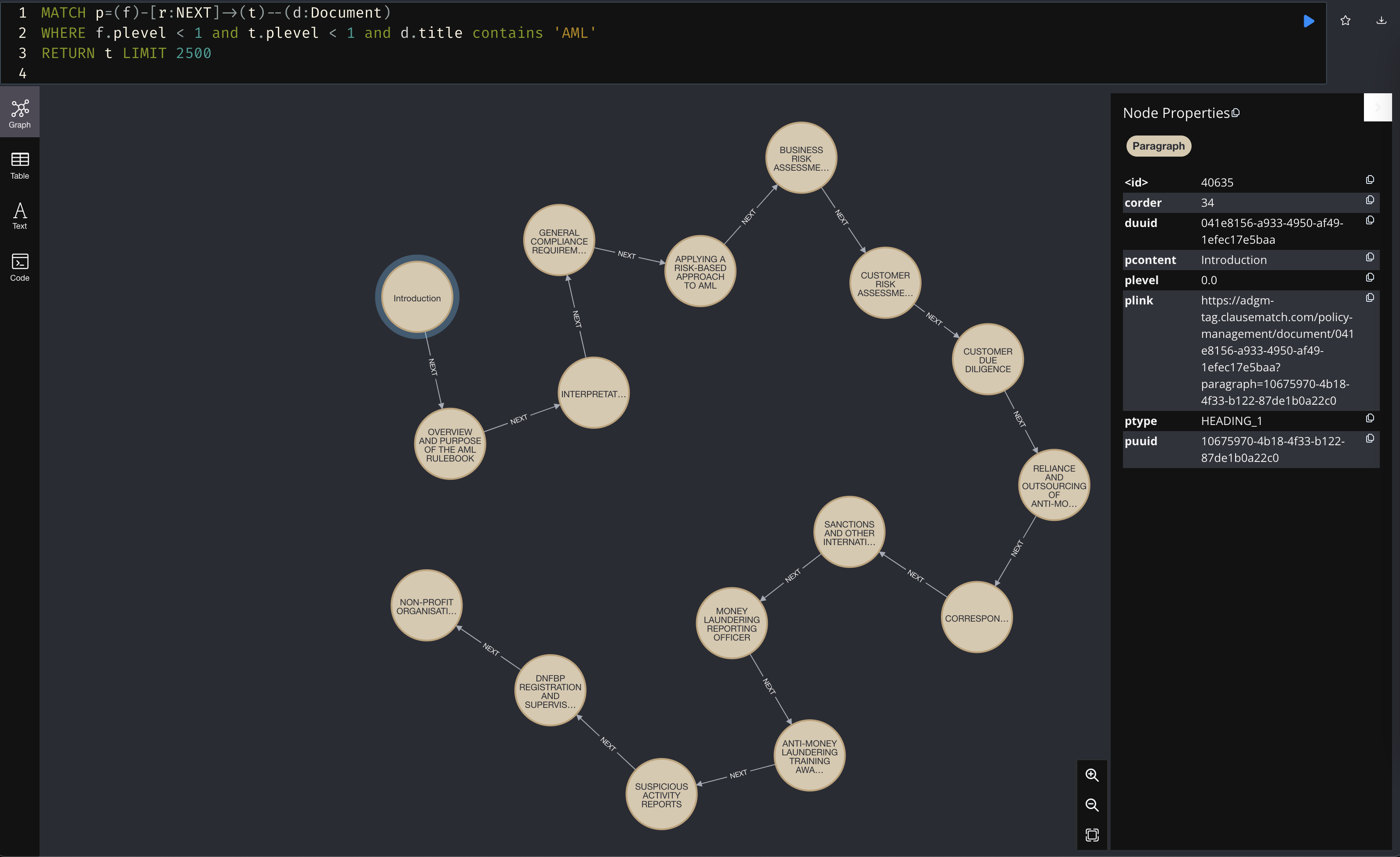}
\end{figure}
\section{Bloom visualization for the exploration of a usage of two permissions in the regulatory documents}
\label{sec:bloom}
\begin{verbatim}
	MATCH (t:Tag) -[r1]- (to:TagOccur) -[r2]- (p:Paragraph) -[r3] - (d:Document)
	WHERE t.ttype in ['PERM'] and t.lemma = 'operate crypto asset business'
	RETURN t, to, p,d, r1, r2, r3
\end{verbatim}
\begin{verbatim}
	MATCH (t:Tag) -[r1]- (to:TagOccur) -[r2]- (p:Paragraph) -[r3] - (d:Document)
	WHERE t.ttype in ['PERM'] and t.lemma contains 'invest' and 
	      t.lemma contains 'princ' and t.lemma contains 'deal'
	RETURN t, to, p,d, r1, r2, r3
\end{verbatim}
\begin{figure}[!htb]
	\centering
	\includegraphics[scale=0.18]{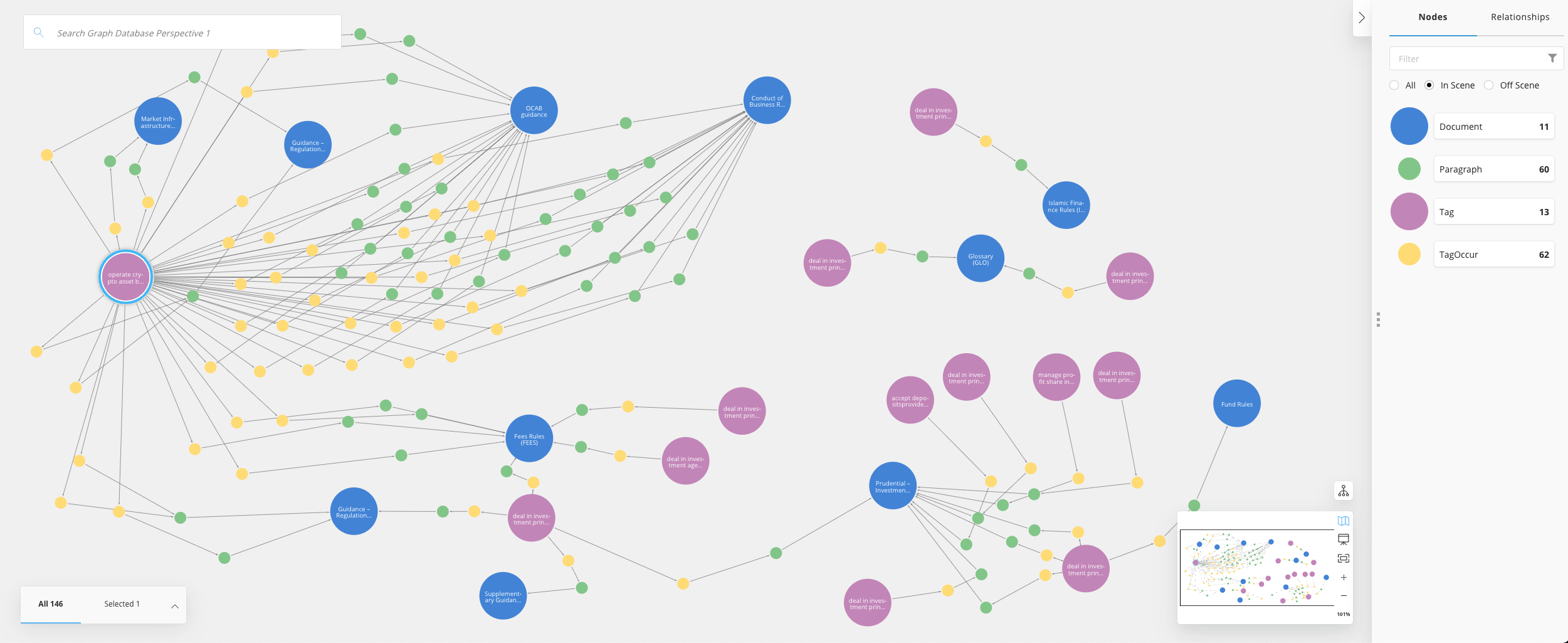}
\end{figure}
\pagebreak
\section{Shortest path between insurance and rule related entities.  Documents(green), concepts(blue), paragraphs(tawny). Neo4j interactive visualization formed with Cypher}
\label{fig:path}	
\begin{figure}[!htb]
	\centering
	\fbox{\includegraphics[scale=0.35]{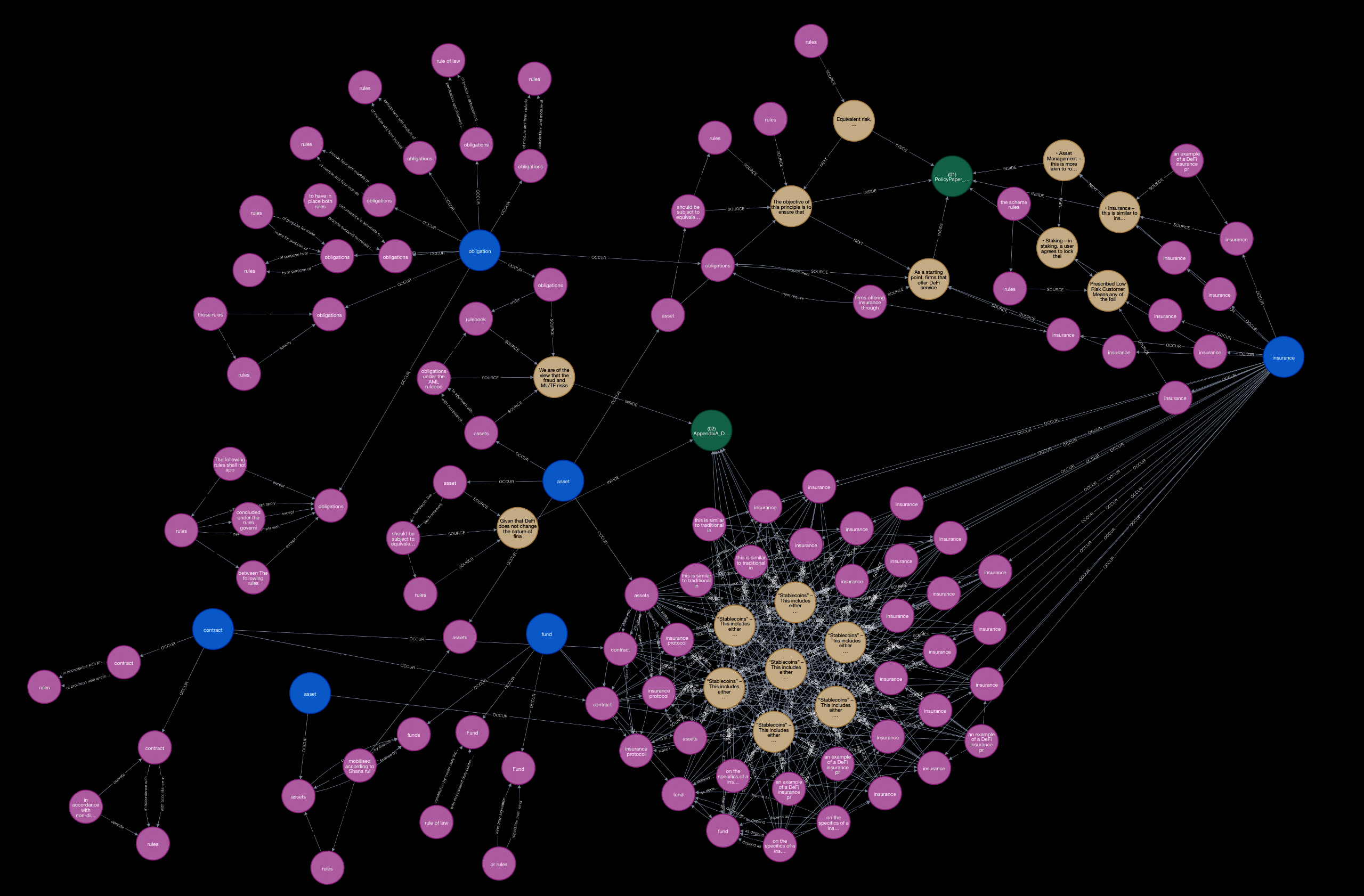}}
\end{figure}
\begin{verbatim}
	MATCH (ent:TagOccur),(mit:TagOccur), p = shortestPath((ent)-[*..4]-(mit)) 	
	WHERE ent.text contains 'insur' AND mit.text contains 'rule'
	RETURN p LIMIT 250
\end{verbatim}
\section{Snippet of tagging issues identified and cleaned manually from the graph}
\label{sec:app_artif}
\begin{table}[tbh!]
	\centering
	\begin{minipage}{0.6\textwidth}
		\begin{verbatim}
			MATCH (t:Tag) -- (to:TagOccur)
			where SIZE(t.lemma) <= 1
			RETURN DISTINCT(to.text), count(*) as cnt
			ORDER BY cnt DESC LIMIT 20
		\end{verbatim}		
		\paragraph{}
		\begin{verbatim}
			MATCH (t:Tag) -- (to:TagOccur)
			where SIZE(t.lemma) <= 0
			DETACH DELETE to, t
			# Deleted 1963 nodes, deleted 17115 relationships, 
			# completed after 253 ms.
		\end{verbatim}	
	\end{minipage}\hfill
	\begin{minipage}{0.4\textwidth}
			\small
			\begin{tabular}{ l 
					l}
					\toprule
					"(Tag text)"&counts  \\ \hline
					\midrule
					"unless"&75 \\
					"if the"&49 \\
					"If the"&43 \\
					"U.A.E."&41 \\
					"whether"&37 \\
					"doing"&37 \\
					"do not"&31 \\
					"whether the"&28 \\
					"U.A.E"&25 \\
					"c)"&25 \\
					"AT1"&24 \\
					"which"&21 \\
					"does not"&21 \\
					")"&18 \\
					"1"&16 \\
					"2"&16 \\
					"7"&16 \\
			\end{tabular}
	\end{minipage}\hfill
\end{table}	
\pagebreak
\section{Tagging depends on the context and is not a direct word matching: ACT example}
\label{sec:appx2}
\begin{verbatim}
[51.58%]  	= 	163/	316  times labeled for 		 Regulated Activity 
[44.14%]  	= 	98/	222  times labeled for 		 Shares 
[55.45%]  	= 	61/	110  times labeled for 		 Contracts of Insurance 
[29.13%]  	= 	30/	103  times labeled for 		 deposits 
[54.35%]  	= 	25/	46  times labeled for 		 Options 
[63.16%]  	= 	24/	38  times labeled for 		 Futures 
[88.89%]  	= 	16/	18  times labeled for 		 sukuk 
[82.35%]  	= 	14/	17  times labeled for 		 joint venture 
[6.83%]  	= 	11/	161  times labeled for 		 Undertaking 
[90.0%]  	= 	9/	10  times labeled for 		 Units in a Collective Investment Fund 
[88.89%]  	= 	8/	9  times labeled for 		 Service-based 
[1.92%]  	= 	8/	417  times labeled for 		 service 
[100.0%]  	= 	8/	8  times labeled for 		 credit agreement 
[2.82%]  	= 	8/	284  times labeled for 		 carrying on, in or from 
[5.04%]  	= 	6/	119  times labeled for 		 held by 
[27.27%]  	= 	6/	22  times labeled for 		 Managing Assets 
[2.15%]  	= 	5/	233  times labeled for 		 marketing 
[1.0%]  	= 	5/	501  times labeled for 		 activity 
[4.76%]  	= 	4/	84  times labeled for 		 provision of 
[30.77%]  	= 	4/	13  times labeled for 		 business activities 
[16.67%]  	= 	3/	18  times labeled for 		 holds or controls 
[21.43%]  	= 	3/	14  times labeled for 		 provision of a service to a Client 
[2.14%]  	= 	3/	140  times labeled for 		 controlled 
[27.27%]  	= 	3/	11  times labeled for 		 intends to carry on 
[60.0%]  	= 	3/	5  times labeled for 		 marketing activities 
[2.68%]  	= 	3/	112  times labeled for 		 can be conducted 
[9.52%]  	= 	2/	21  times labeled for 		 involves provision 
[1.96%]  	= 	2/	102  times labeled for 		 carrying on a Regulated Activity 
[33.33%]  	= 	2/	6  times labeled for 		 held or controlled 
[100.0%]  	= 	2/	2  times labeled for 		 develop or to undertake 
[1.53%]  	= 	2/	131  times labeled for 		 offered in 
[0.49%]  	= 	2/	412  times labeled for 		 is set up by 
[11.76%]  	= 	2/	17  times labeled for 		 provision of a service 
[0.15%]  	= 	1/	684  times labeled for 		 invest 
[2.13%]  	= 	1/	47  times labeled for 		 In the calculation 
[1.18%]  	= 	1/	85  times labeled for 		 received 
[50.0%]  	= 	1/	2  times labeled for 		 provided for the purposes 
[0.26%]  	= 	1/	384  times labeled for 		 Fund Manager 
[100.0%]  	= 	1/	1  times labeled for 		 Retail authorisation 
[16.67%]  	= 	1/	6  times labeled for 		 expertise 
[0.08%]  	= 	1/	1257  times labeled for 		 acting as 
[11.11%]  	= 	1/	9  times labeled for 		 carrying on an activity 
[50.0%]  	= 	1/	2  times labeled for 		 directly held 
[25.0%]  	= 	1/	4  times labeled for 		 must not carry on 
[33.33%]  	= 	1/	3  times labeled for 		 due and payable 
[2.86%]  	= 	1/	35  times labeled for 		 engages with 
[33.33%]  	= 	1/	3  times labeled for 		 advising or arranging 
[50.0%]  	= 	1/	2  times labeled for 		 contracts for differences 
[11.11%]  	= 	1/	9  times labeled for 		 carries on or intends to carry on 
[100.0%]  	= 	1/	1  times labeled for 		 made payable 
[100.0%]  	= 	1/	1  times labeled for 		 Large Undertaking 
[100.0%]  	= 	1/	1  times labeled for 		 giving and receiving of instructions 
[3.85%]  	= 	1/	26  times labeled for 		 demonstrate 
[16.67%]  	= 	1/	6  times labeled for 		 dedicated to 
[33.33%]  	= 	1/	3  times labeled for 		 held directly or indirectly 
[100.0%]  	= 	1/	1  times labeled for 		 promotional activities 
[100.0%]  	= 	1/	1  times labeled for 		 when it first carries on 
[100.0%]  	= 	1/	1  times labeled for 		 operated in accordance with the instructions 
[4.55%]  	= 	1/	22  times labeled for 		 contribute 
[100.0%]  	= 	1/	1  times labeled for 		 regard to its engagement 
[1.85%]  	= 	1/	54  times labeled for 		 participated 
[100.0%]  	= 	1/	1  times labeled for 		 Certificates representing certain Financial Instruments 
[100.0%]  	= 	1/	1  times labeled for 		 carried on, or held out as being carried on 
[25.0%]  	= 	1/	4  times labeled for 		 business purposes 
[100.0%]  	= 	1/	1  times labeled for 		 at the early stages of interaction 
[0.47%]  	= 	1/	211  times labeled for 		 arrangements 
[100.0%]  	= 	1/	1  times labeled for 		 inclined to act in accordance with the instructions 
[100.0%]  	= 	1/	1  times labeled for 		 Advisory and arranging 
[33.33%]  	= 	1/	3  times labeled for 		 communicating information 
\end{verbatim}
\end{document}